\pdfoutput=1
\PassOptionsToPackage{prologue,dvipsnames}{xcolor}
\documentclass[11pt]{article}
\usepackage[table]{xcolor}
\usepackage[preprint]{acl}

\usepackage{times}
\usepackage{latexsym}

\usepackage[T1]{fontenc}

\usepackage[utf8]{inputenc}

\usepackage{microtype}

\usepackage{inconsolata}

\usepackage{graphicx}

\usepackage{algorithm}
\usepackage[noend]{algpseudocode}
\usepackage{amsfonts}
\usepackage{amsmath}
\usepackage{booktabs}
\usepackage{caption}
\usepackage{cleveref}
\usepackage{enumitem}
\usepackage{float}
\usepackage{graphicx}
\usepackage{hyperref}
\usepackage{makecell}
\usepackage{multirow}
\usepackage{natbib}

\usepackage{wrapfig}
\usepackage{pifont}
\usepackage{times}
\usepackage{tcolorbox}
\usepackage{latexsym}
\definecolor{cvprblue}{rgb}{0.21,0.49,0.74}

\usepackage{tikz}
\usepackage{xcolor}

\definecolor{pastelblue}{RGB}{135,180,225}    
\definecolor{pastelgreen}{RGB}{120,200,150}    
\definecolor{pastelred}{RGB}{240,130,140}      
\definecolor{pastelpurple}{RGB}{180,160,210}    
\definecolor{pastelorange}{RGB}{255,180,90}     
\definecolor{pastelyellow}{RGB}{250,220,120}    
\definecolor{pastelteal}{RGB}{100,190,190}    

\newcommand{\ralign}{{\textsc{Re-Align}}}
\newcommand\semiLarge{\@setfontsize\semiLarge{13.15}{15.18}}
\newcommand{\realignlogo}{\raisebox{-0.5em}
{\includegraphics[height=2.5em]{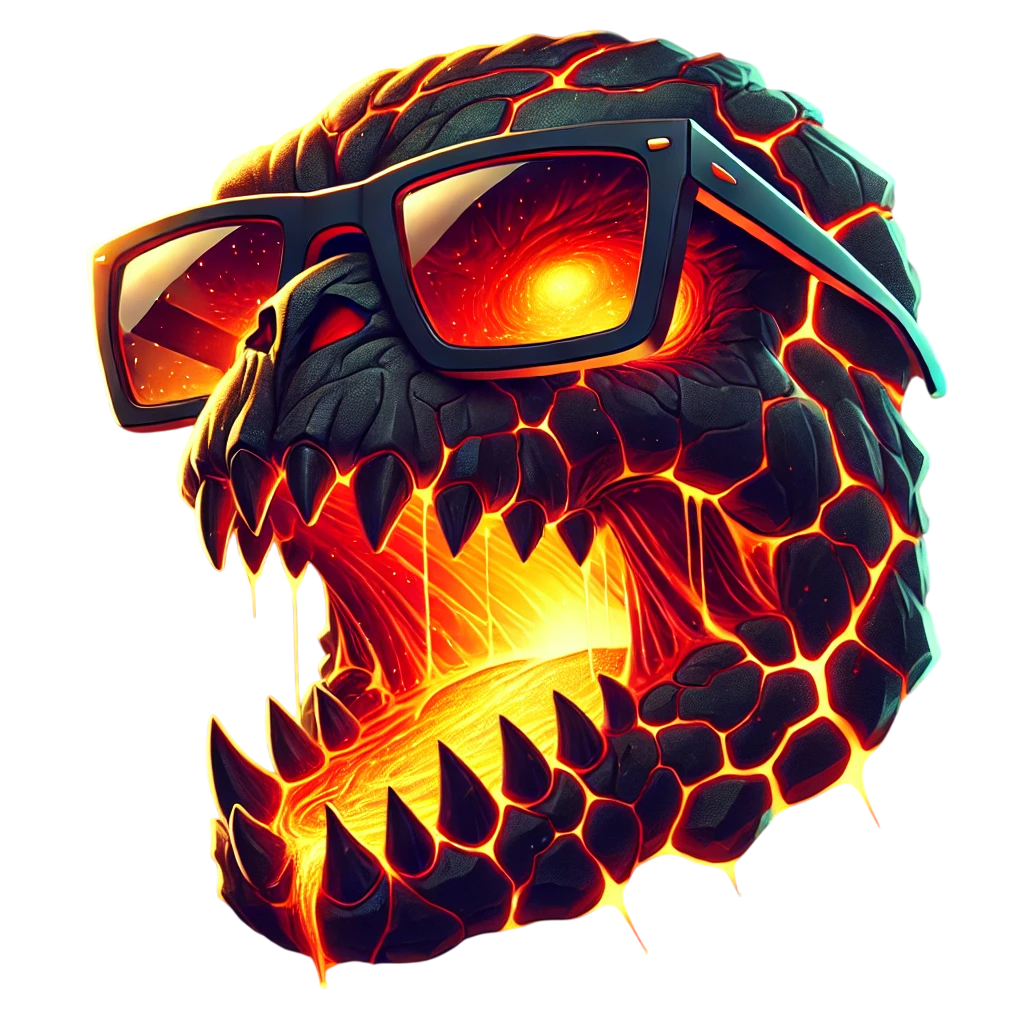}}}

\title{\realignlogo \ \textsc{Re-Align}: Aligning Vision Language Models via Retrieval-Augmented Direct Preference Optimization}

\author{
Shuo Xing$^1$,\quad 
Peiran Li$^1$,\quad
Yuping Wang$^2$,\quad
Ruizheng Bai$^1$,\quad 
Yueqi Wang$^3$,\quad \\
\bf Chan-Wei Hu$^1$,\quad
Chengxuan Qian$^1$,\quad
Huaxiu Yao$^4$,\quad
Zhengzhong Tu$^1$\thanks{\  Corresponding author.}\\
\\
$^1$Texas A\&M University \quad
$^2$University of Michigan \quad
$^3$UIUC \quad 
$^4$UNC Chapel Hill \\
\\
\texttt{\{shuoxing,tzz\}@tamu.edu}
}

\begin{document}
\maketitle
\begin{abstract}
The emergence of large Vision Language Models (VLMs) has broadened the scope and capabilities of single-modal Large Language Models (LLMs) by integrating visual modalities, thereby unlocking transformative cross-modal applications in a variety of real-world scenarios.
Despite their impressive performance, VLMs are prone to significant hallucinations, particularly in the form of cross-modal inconsistencies. 
Building on the success of Reinforcement Learning from Human Feedback (RLHF) in aligning LLMs, recent advancements have focused on applying direct preference optimization (DPO) on carefully curated datasets to mitigate these issues. 
Yet, such approaches typically introduce preference signals in a brute-force manner, neglecting the crucial role of visual information in the alignment process. 
In this paper, we introduce \ralign{}, a novel alignment framework that leverages image retrieval to construct a dual-preference dataset, effectively incorporating both textual and visual preference signals.
We further introduce rDPO, an extension of the standard direct preference optimization that incorporates an additional visual preference objective during fine-tuning. 
Our experimental results demonstrate that \ralign{} not only mitigates hallucinations more effectively than previous methods but also yields significant performance gains in general visual question-answering (VQA) tasks. Moreover, we show that \ralign{} maintains robustness and scalability across a wide range of VLM sizes and architectures.
This work represents a significant step forward in aligning multimodal LLMs, paving the way for more reliable and effective cross-modal applications.
\end{abstract}

\begin{figure}[htbp]
    \centering
    \includegraphics[width=1.\linewidth]{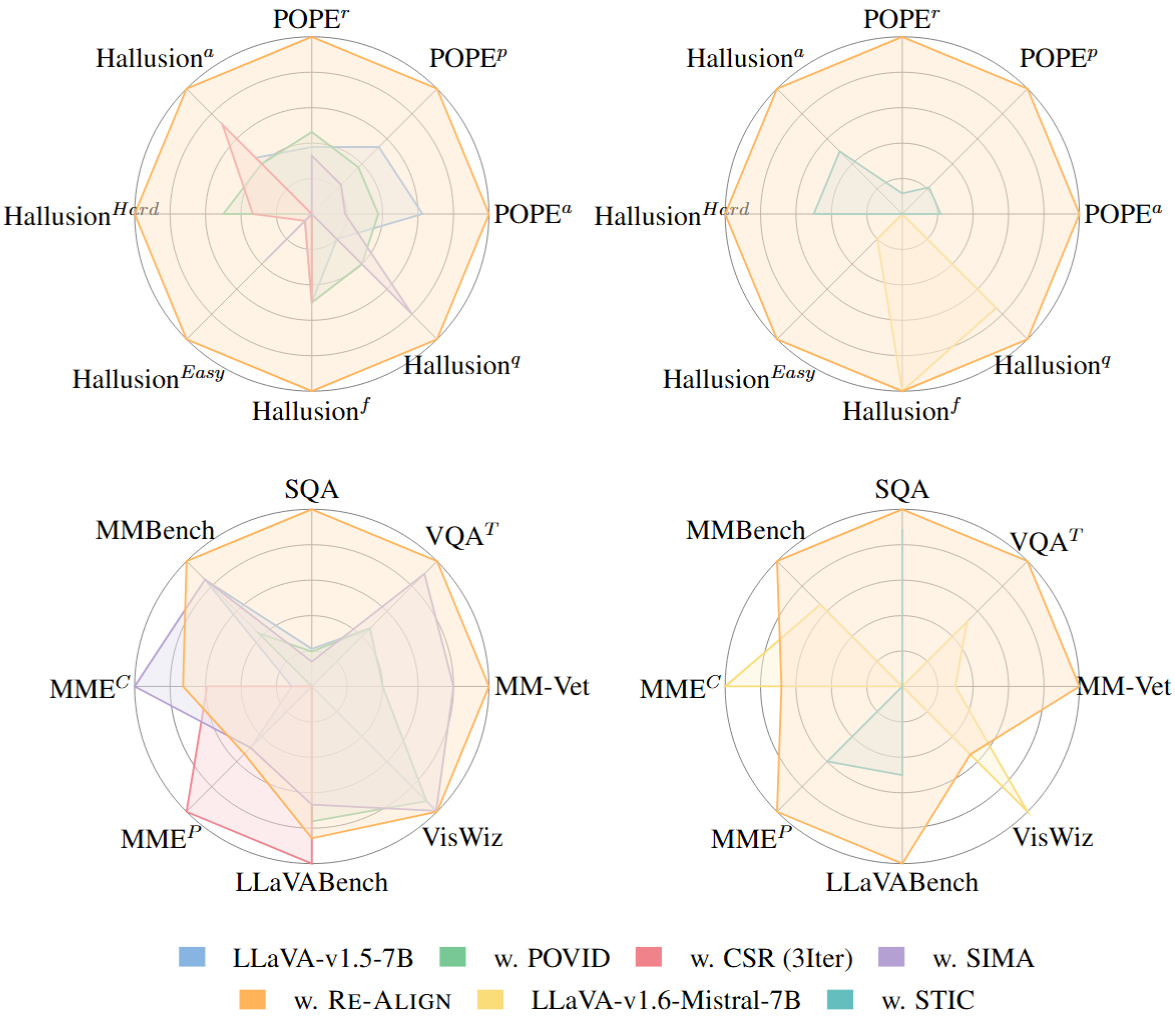}
    \caption{Benchmark performance comparison (min-max normalized).}
    \label{fig:radar}   
    \vspace{-5mm}
\end{figure}

\section{Introduction}
The recent emergence of powerful Vision Language Models (VLMs)~\citep{li2022blip, li2023blip2, liu2024llava,llavanext,llama3.2, Qwen-VL, Qwen2VL, lu2024deepseek,wu2024deepseek,bai2025qwen2.5vl,fan2025vlm3rvisionlanguagemodelsaugmented} has significantly extended the capabilities of Large Language Models (LLMs)~\citep{devlin2018bert, radford2019gpt2,brown2020gpt3,team2023gemini,roziere2023codellama,touvron2023llama,touvron2023llama2, raffel2020t5,qwen2,qwen2.5} into the visual domain, paving the way for innovative real-world applications that integrate multimodal information~\citep{moor2023med,li2024llava-med, shao2024lmdrive,openemma,rana2023sayplan,kim2024openvla}. Despite their promising performance, VLMs remain susceptible to hallucinations—instances where the model produces outputs containing inaccurate or fabricated details about objects, attributes, and the logical relationships inherent in the input image~\citep{rohrbach2018object,bai2024hallucination}. Several factors contribute to this cross-modal inconsistency, including the separate low-quality or biased training data, imbalanced model architectures, and the disjoint pretraining of the vision encoder and LLM-backbone~\citep{cui2023holistic,bai2024hallucination,zhou2024povid}. 

To mitigate the hallucinations in VLMs, the Directed Preference Optimization (DPO) techniques have been widely adopted~\citep{deng2024stic,zhou2024povid,fang2024vila,zhou2024calibrated,guo2024direct,chen2024dress,wang2024enhancing,yu2024rlhf,li2023silkie,wang2024mdpo,xiao2025detecting,xie2024v,fu2024mitigating}. This involves constructing datasets enriched with human preference signals specifically targeting hallucinations, and then fine-tuning the models using algorithms like Direct Preference Optimization (DPO)~\citep{rafailov2024direct}. Existing methods generate the preference data by perturbing the ground truth responses~\citep{zhou2024povid} and corrupting the visual inputs/embeddings~\citep{deng2024stic, amirloo2024understanding} to generate rejected responses or correcting/refining responses to produce chosen responses~\citep{chen2024dress,yu2023reformulating}. While methods based on response refinement yield the most reliable preference signals, they face scalability challenges due to the significant costs of manual correction processes. Conversely, directly corrupting input visual information or ground truth responses is overly simplistic, as this brute-force approach fails to generate plausible and natural hallucinations in a controlled manner. Moreover, during fine-tuning, directly applying DPO may cause the model to overly prioritize language-specific preferences, which potentially leads to suboptimal performance and an increased propensity for hallucinations~\citep{wang2024mdpo}.

In this paper, we propose \textbf{\ralign{}}, a novel framework that alleviates VLM hallucinations by integrating {image retrieval} with direct preference optimization (DPO). Our method deliberately injects controlled hallucinations into chosen responses using image retrieval, generating rejected responses that offer more plausible and natural preference signals regarding hallucinations. Additionally, by incorporating both the retrieved image and the original input image, \ralign{} constructs a {dual preference dataset}. This dataset is then leveraged to finetune VLMs with our proposed \textbf{rDPO} objective—an extension of DPO that includes an additional visual preference optimization objective, further enhancing the alignment process with valuable {visual preference signals}.

\section{Preliminaries}
To mitigate hallucinations in VLMs, we introduce an alignment framework based on direct preference optimization (DPO) with image retrieval. In this section, we present preliminary definitions and notations for VLMs and preference optimization, which serve as the foundation for our proposed framework. 

\paragraph{Vision Language Models} 
VLMs typically consist of three main components: a vision encoder $f_v(\cdot)$, a projector $f_p(\cdot)$, and an LLM backbone $\mathcal{L}(\cdot)$. Given a multimodal input query $(x,v)$, where $x$ is a textual instruction and $v$ is a visual image, VLMs generate a corresponding response $y = [y_1, \cdots, y_m]$ autoregressively. Here, each $y_i$ represents an output token, and $m$ denotes the total number of tokens in the generated response.

\paragraph{Direct Preference Learning} Reinforcement Learning from Human Feedback (RLHF) \citep{christiano2017deep, ziegler2019fine} is a key approach for aligning machine learning models with human preferences. Among these techniques, the Direct Preference Optimization (DPO) algorithm~\citep{rafailov2024direct} stands out for its popularity and for demonstrating superior alignment performance. We represent a VLM with a policy $\pi$, which, given an input query $(x,v)$, generates a response $y$ from the distribution $\pi(\cdot|x,v)$. We denote by $\pi_0$ the initial VLM model, fine-tuned on instruction-following VQA data by supervised fine-tuning (SFT). Specifically, we define a preference dataset $\mathcal{D} = \{(x, v, y_w, y_l)\}$, where for each input, the response $y_w$ is preferred to the response $y_l$. The DPO objective is formulated as follows, leveraging the preference dataset $\mathcal{D}$: 
\begin{align*}
    &\mathcal{L}_\text{DPO} = - \mathbb{E}_{(x,v,y_w,y_l) \sim \mathcal{D}}\\
    &\bigg[\log \sigma \bigg(\beta \log \frac{\pi_\theta (y_w|x,v)}{\pi_0 (y_w|x,v)}
    - \beta \log \frac{\pi_\theta (y_l|x,v)}{\pi_0 (y_l|x,v)} \bigg) \bigg].
\end{align*}

Compared to deep RL-based methods like Proximal Policy Optimization (PPO)~\citep{schulman2017proximal, christiano2017deep, ziegler2019fine}, DPO is more computationally efficient, easier to tune, and thus more widely adopted~\citep{dong2024rlhf}. 

\paragraph{Image Retrieval} Image retrieval aims to find relevant images from large databases -- such as vector databases or indexed corpora -- based on semantic similarity criteria~\citep{hu2025mrag}. In this paper, we convert all images into vector representations and utilize the cosine similarity metric to evaluate their proximity to a reference image. The similarity between two images, $v_1$ and $v_2$, is computed as follows:
\begin{align*}
    s = \bigg< \frac{f_p(v_1)}{||f_p(v_1)||}, \frac{f_p(v_2)}{||f_p(v_2)||} \bigg>,
\end{align*}
where $<\cdot, \cdot>$ denotes the inner product in $l_2$ space, $f_p(v_i)$ represents the image embeddings generated by the vision encoder $f_v(\cdot)$ of VLMs. In this paper, we employ the FAISS library~\citep{douze2024faiss,johnson2019billion} for efficient vector searches, retrieving the top-$k$ most relevant images.

\section{Methods}
In this paper, we propose \ralign{}, a novel framework that integrates preference optimization with image retrieval to improve cross-modal alignment in VLMs. 
\begin{figure}[htbp]
    \centering
    \includegraphics[width=1.\linewidth]{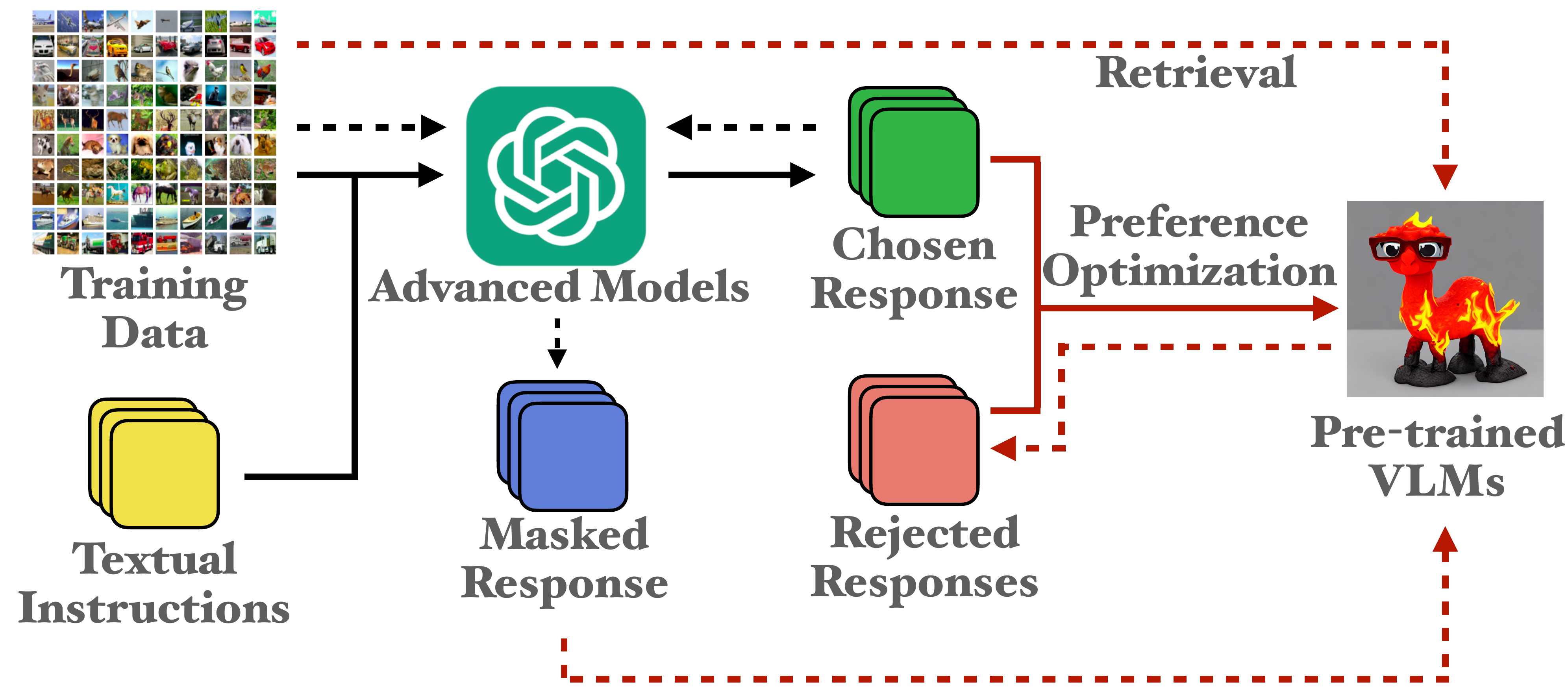}
    \caption{Illustration of \ralign{} framework.}
    \label{fig:harp-pipeline}   
\end{figure}
As shown in Figure \ref{fig:harp-pipeline}, the process begins with an advanced VLM generating chosen responses from input images from the training set. A selective masking process is then applied, strategically omitting segments associated with objects, attributes, or logical relationships identified in the image. Next, leveraging the retrieved image from the same training dataset and the masked responses, the hallucination-prone VLM is prompted to complete the masked elements, obtaining rejected responses. The generated preference pairs (chosen vs. rejected) are then used to fine-tune the VLM with $\mathcal{L}_{\text{rDPO}}$ (\cref{eq:rdpo}), a preference objective that integrates both visual and textual information to penalize hallucinations and reinforce grounded reasoning. Algorithm \ref{alg:harp} in Appendix \ref{app:alg-ralign} provides an overview of \ralign{}, while the detailed process is explained in the following subsections.

\subsection{Preference Generation}
\begin{figure*}[htbp]
    \centering
    \includegraphics[width=0.95\linewidth]{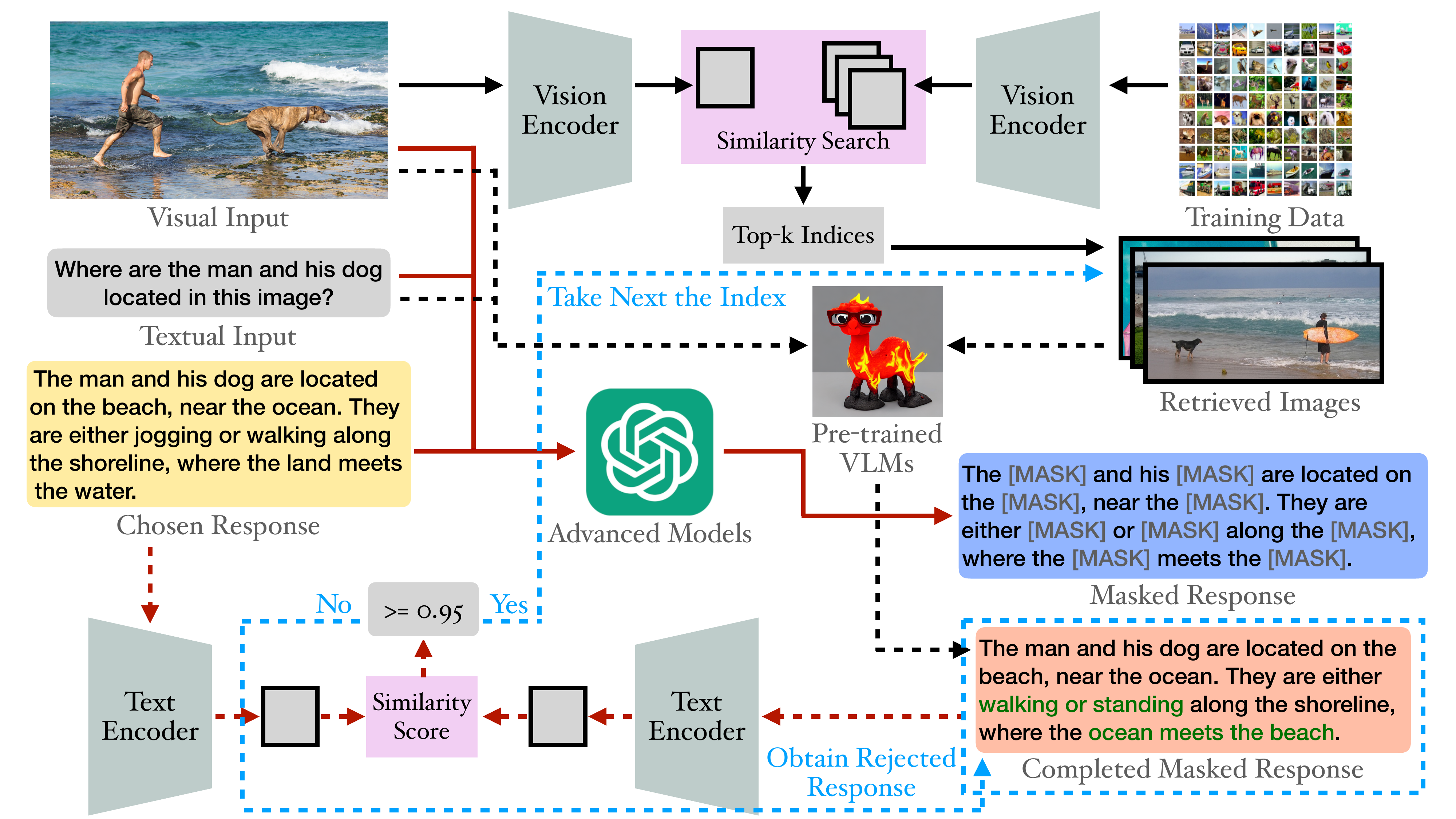}
    \caption{Illustration of the preference generation process, utilizing the original vision encoder from initial VLMs and the SentenceTransformer as the text encoder. }
    \label{fig:gen-reject}   
\end{figure*}
Generating high-quality preference data, which includes both accurate ground-truth responses and controlled hallucinated examples, is crucial for effective preference optimization in pre-trained VLMs.
Existing methods construct preference data by perturbing ground-truth responses~\citep{zhou2024povid}, corrupting visual inputs/embeddings~\citep{deng2024stic, amirloo2024understanding} to create rejected responses, or refining responses to obtain chosen responses~\citep{chen2024dress,yu2023reformulating}. Refinement produces high-quality preference data but comes at a high cost, whereas direct corruption is more scalable yet tends to generate unrealistic hallucinations and fails to produce plausible, natural ones in a controlled manner. To address these limitations, we introduce a novel image retrieval-based pipeline for preference data construction as shown in Figure \ref{fig:gen-reject}, which consists of three key stages:
\begin{itemize}[leftmargin=*,nosep]
    \item \textbf{Strategical masking:} Given an input pair $(x_i,v_i)$ and its corresponding chosen response $y_w$ generated by a pretrained VLM, a strategic masking process removes words or segments associated with objects, attributes, or logical relationships inferred from the image, producing the masked response $y_m$.
    \item \textbf{Image retrieval:} All images $\{v_i\}$ in the training set are embedded using the original vision encoder of the pre-trained VLMs, forming the knowledge base $\mathcal{K}$. The top-$k$ most similar images to $v_i$ are then retrieved from $\mathcal{K}$ using a cosine similarity search.
    \item \textbf{Inducing hallucinations:} VLMs are prompted to generate a candidate completion $y_m$ for the masked response conditioned on the instruction $x$ and a retrieved image $v_{j_t}$ where $t \in [1,k]$ denotes the rank of images based on their cosine similarity to the input $v_i$. Both the chosen response $y_w$ and the reconstructed response $y_c$ are embedded using a $\text{SentenceTransformer}$ model. If the cosine similarity between these embeddings falls below $0.95$, $y_c$ is designated as the rejected response $y_l$. Otherwise, the process continues with the next image $v_{j_{t+1}}$ in the similarity-ranked sequence until a suitable candidate is identified or all $k$ retrieved images have been examined. 
    
\end{itemize}

\subsection{Preference Optimization}
The curated preference dataset is subsequently used to fine-tune VLMs through direct preference learning. We propose retrieval-augmented direct preference optimization (rDPO), an extension of DPO that integrates an additional visual preference optimization objective. Given a preference dataset $\mathcal{D} = \{x,v,v_l,y_w,y_l\}$, the retrieval-augmented direct preference optimization objective is formulated as follows:
\begin{align*}
    &\mathcal{L}_{\text{vDPO}} = - \mathbb{E}_{(x,v,v_l,y_w,y_l) \sim \mathcal{D}}\\
    &\bigg[\log \sigma \bigg(\beta \log \frac{\pi_\theta (y_w|x,v)}{\pi_0 (y_w|x,v)}
    - \beta \log \frac{\pi_\theta (y_w|x,v_l)}{\pi_0 (y_w|x,v_l)} \bigg) \bigg],
\end{align*}
where $(x,v)$ denotes the input query of VLMs, $(y_w,y_l)$ represents the preference responses pair, and $v_l$ is the retrieved image for $v$. The loss function of rDPO is the combination of standard DPO objective and visual preference optimization:
\begin{align}\label{eq:rdpo}
    \mathcal{L}_{\text{rDPO}} = \mathcal{L}_{\text{DPO}} + \mathcal{L}_{\text{vDPO}}.
\end{align}
By incorporating both textual and visual preference signals, our approach allows VLMs to effectively exploit multimodal information during optimization, in contrast to prior alignment methods that depend exclusively on language-based preferences. In contrast to mDPO~\citep{wang2024mdpo}, which introduces image preference by randomly cropping the original input images, rDPO adopts retrieval-augmented generation to integrate visual preference signals in a more coherent and semantically meaningful way.

\begin{table*}[htbp]
  \footnotesize
  \setlength{\tabcolsep}{5pt}
  \begin{center}
    \begin{tabular}{lcccccccccccccccccccr}
      \toprule

      Methods
      & POPE$^r$
      & POPE$^p$
      & POPE$^a$
      & Hallusion$^q$
      & Hallusion$^f$
      & Hallusion$^{Easy}$
      & Hallusion$^{Hard}$
      & Hallusion$^a$
      \\
      \midrule

      LLaVA-v1.5-7B
      & 88.14 
      & 87.23
      & 85.10
      & 10.3297
      & 18.2081
      & 41.7582
      & 40.2326
      & 46.3242
      \\

      \textbf{w.} LLaVA-RLHF
      & 84.77	
      & 84.60	
      & 83.40
      & 10.2859
      & 18.7861
      & 38.2418
      & 40.6744
      & 44.6528
      \\

      \textbf{w.} POVID
      & 88.21 
      & 87.16
      & 85.06
      & 10.5495
      & 18.2081
      & 41.5385
      & 40.9302
      & 46.6785
      \\

      \textbf{w.} CSR (3Iter)
      & 87.83
      & 87.00
      & 85.00
      & 10.1099 
      & 18.2081 
      & 41.7582
      & 40.6977
      & 46.9442 
      \\

      \textbf{w.} SIMA
      & 88.10
      & 87.10
      & 85.03
      & 10.9890 
      & 17.6301
      & 43.0549
      & 40.2326
      & 45.2728
      \\

      \textbf{w.} mDPO
      & 88.17 
      & 87.13
      & 85.03
      & 9.8901 
      & 18.4971
      & 41.978 
      & 40.000
      & 46.1470 
      \\

      \rowcolor[gray]{0.9} \textbf{w.} \ralign{}
      & \textbf{88.65}
      & \textbf{87.43}
      & \textbf{85.16}
      & \textbf{11.2088}
      & \textbf{18.7861}
      & \textbf{45.5165}
      & \textbf{41.6279}
      & \textbf{47.6156}
      \\

      \midrule

      \makecell[l]{LLaVA-v1.6-\\ \quad \quad Mistral-7B} 
      & 88.83
      & 87.93
      & 86.43 
      & 13.6264 
      & 19.0751 
      & 47.4725 
      & 33.4884 
      & 46.0585 
      \\

      \textbf{w.} STIC
      & 89.03
      & 88.20
      & 86.56 
      & 12.9670
      & 17.3410 
      & 47.2527 
      & 34.1860 
      & 46.3242 
      \\

      \rowcolor[gray]{0.9} \textbf{w.} \ralign{}
      & \textbf{90.55} 
      & \textbf{89.20}
      & \textbf{87.03}
      & \textbf{13.8462} 
      & \textbf{19.0751}
      & \textbf{48.3516}
      & \textbf{34.8837} 
      & \textbf{46.5899} 
      \\

      \bottomrule
    \end{tabular}
  \end{center}
  \caption{Impact of \ralign{} across hallucination benchmarks for VLMs, and comparisons with baselines. }
  \label{tab:hallucination-task}
\end{table*}

\begin{table*}[htbp]
  \footnotesize
  \setlength{\tabcolsep}{5pt}
  \begin{center}
    \begin{tabular}{lccccccccc}
      \toprule

      Methods
      & SQA
      & TextVQA
      & MM-Vet
      & VisWiz
      & LLaVABench
      & MME$^P$
      & MME$^C$
      & MMBench
      & Avg. Rank
      \\
      \midrule

      LLaVA-v1.5-7B
      & 66.02
      & 58.18
      & 31.6
      & 50.03
      & 64.1
      & 1510.28
      & 357.85
      & \underline{64.60}
      & 3.875
      \\

      \textbf{w.} LLaVA-RLHF
      & 63.11	
      & 56.89
      & 31.8
      & 49.57
      & 60.2
      & 1378.90
      & 282.85
      & 64.39
      & 6
      \\

      \textbf{w.} POVID
      & 65.98
      & 58.18
      & 31.8
      & 49.80
      & 67.3
      & 1495.91
      & 356.07
      & 64.34
      & 4.375
      \\

      \textbf{w.} CSR (3Iter)
      & 65.46
      & 57.86
      & 31.6
      & 47.02 
      & \textbf{68.3}
      & \textbf{1525.44}
      & 365.35
      & 64.08
      & 4.5
      \\

      \textbf{w.} SIMA
      & 65.83
      & \underline{58.48}
      & \underline{32.0} 
      & \underline{50.04}
      & 66.9
      & 1510.33
      & \textbf{371.78} 
      & 64.60 
      & \underline{2.75}
      \\

      \textbf{w.} mDPO
      & \underline{67.53}
      & 57.90
      & 31.3
      & 50.04
      & 59.0
      & 1510.74
      & 335.71
      & 64.60
      & 4.25
      \\

      \rowcolor[gray]{0.9} \textbf{w.} \ralign{}
      & \textbf{68.10}
      & \textbf{58.55}
      & \textbf{32.1}
      & \textbf{50.06}
      & \underline{67.7}
      & \underline{1511.79}
      & \underline{367.50}
      & \textbf{64.69}
      & \textbf{1.375}
      \\

      \midrule

      \makecell[l]{LLaVA-v1.6-\\ \quad \quad Mistral-7B}  
      & 76.02
      & \underline{63.80}
      & \underline{47.6}
      & \textbf{59.85}
      & 80.2
      & 1494.22
      & \textbf{323.92}
      & \underline{69.33}
      & \underline{2.125}
      \\

      \textbf{w.} STIC
      & \underline{76.42}
      & 63.50
      & 47.3
      & 54.21
      & \underline{81.0}
      & \underline{1504.91} 
      & 308.21 
      & 69.16 
      & 2.625
      \\

      \rowcolor[gray]{0.9} \textbf{w.} \ralign{}
      & \textbf{76.47}
      & \textbf{64.08}
      & \textbf{48.3}
      & \underline{57.27} 
      & \textbf{81.8} 
      & \textbf{1512.09} 
      & \underline{318.93} 
      & \textbf{69.42}
      & \textbf{1.25}
      \\

      \bottomrule
    \end{tabular}
  \end{center}
  \caption{Impact of \ralign{} across general benchmarks for VLMs, and comparisons with baselines. }
  \label{tab:general-task}
\end{table*}

\section{Experiments}
We conduct three categories of experiments to empirically validate the effectiveness of our proposed method. First, we evaluate the ability of \ralign{} to mitigate hallucinations and improve generalizability across diverse VQA tasks, demonstrating its consistent superiority over baseline approaches and achieving state-of-the-art performance. Next, we examine \ralign{}'s effectiveness in aligning VLMs across various model sizes and architectures, including both text-to-image and unified models, where it delivers substantial performance over vanilla models and existing baselines. Finally, we assess the impact of our proposed rDPO objective in preference optimization, showing that it consistently surpasses standard DPO in aligning VLMs and achieving superior results in both halluciation mitigation and general tasks.


\subsection{\ralign{} for VLMs Alignment}
\paragraph{Datasets} We conducted experiments on both hallucination detection and general VQA tasks. Specifically, we assess our method’s performance in hallucination detection using the POPE dataset~\citep{pope} and HallusionBench~\citep{guan2023hallusionbench}. For general VQA tasks, we leverage a diverse suite of benchmarks including ScienceQA~\citep{lu2022scienqa}, TextVQA~\citep{singh2019towards}, MM-Vet~\citep{yu2023mm}, VisWiz~\citep{gurari2018vizwiz}, LLaVABench~\citep{llavabench}, MME~\citep{Fu2023mme}, and MMBench~\citep{liu2024mmbench}.

\paragraph{Beslines} We compare our method with several widely adopted alignment frameworks for VLMs, including \textbf{LLaVA-RLHF}~\citep{sun2023aligning}, \textbf{POVID}~\cite{zhou2024povid}, \textbf{CSR}~\citep{zhou2024calibrated}, \textbf{SIMA}~\citep{wang2024enhancing}, \textbf{STIC}~\citep{deng2024stic}. For more details on these baselines, please refer to the Appendix.

\paragraph{Experimental Setup} We sample 11k images from the LLaVA-Instruct-150K dataset~\citep{liu2024llava} to construct preference data, as illustrated in Figure \ref{fig:gen-reject}. These images are initially used to generate QA pairs based on image captions and simple VQA tasks using GPT-4o mini~\citep{gpt4omini}. Furthermore, the images are encoded using \texttt{clip-vit-large-patch14}~\citep{radford2021learning} to construct the knowledge base for image retrieval. For rejected responses, we use GPT-4o mini to mask the chosen response and \texttt{all-mpnet-base-v2}~\citep{reimers-2019-sentence-bert} to compute the similarity between the completed masked response and the original chosen response. We use LLaVA-v1.5-7B~\citep{liu2024llava} and LLaVA-v1.6-Mistral-7B~\citep{llavanext} as our backbone models and perform \ralign{} fine-tuning for $1$ epoch. All evaluations are conducted with a temperature setting of 0, and baseline results are reproduced using open-sourced model weights.

\paragraph{Results} Table \ref{tab:hallucination-task} shows the performance of \ralign{} compared to baseline methods on hallucination benchmarks. Notably, \ralign{} achieves the best among the evaluated methods on both POPE and HallusionBench for LLaVA-v1.5-7B~\citep{liu2024llava} and LLaVA-v1.6-Mistral-7B~\citep{llavanext}, highlighting the effectiveness of our approach in mitigating hallucinations of VLMs. As shown in Table \ref{tab:general-task}, \ralign{} can provide generally on-par or better performance than the vanilla models and baseline alignment methods on each evaluated general VQA task, ultimately achieving the best overall results. This finding indicates that \ralign{} can enhance hallucination mitigation without compromising general performance.

\subsection{Scalability and Generalizability} 
\paragraph{Experimental Setup} The experimental setup follows the same setting as VLMs alignment experiments, except for the backbone models, where we employ a diverse array of VLMs varying in size and architecture:
\begin{itemize}[leftmargin=*,nosep]
    \item \textbf{Image-to-Text models}: the typical architecture of VLMs, where a vision encoder is integrated with an LLM to enable cross-modal understanding. In this section, we evaluate \ralign{} on LLaVA-v1.5-7B~\citep{liu2024llava}, LLaVA-v1.5-13B~\citep{liu2024llava}, LLaVA-v1.6-Vicuna-7B~\citep{llavanext}, LLaVA-v1.6-Vicuna-13B~\citep{llavanext}, Qwen2.5-VL-3B-Instruct~\citep{bai2025qwen2.5vl}, and Qwen2.5-VL-7B-Instruct~\citep{bai2025qwen2.5vl}.
    
    \item \textbf{Unified Models}: encoder-decoder architecture that decouples visual encoding for multimodal understanding and generation. We evaluate \ralign{} on Janus-Pro-1B~\citep{chen2025janus} and Janus-Pro-7B~\citep{chen2025janus}.
\end{itemize}

\begin{table}[htbp]
  \footnotesize
  \begin{center}
    \begin{tabular}{llllllllllllllll}
      \toprule

      Methods
      & POPE$^r$
      & POPE$^p$
      & POPE$^a$
      \\
      \midrule

      Janus-Pro-1B
      & 85.46
      & 85.03
      & 84.13 
      \\

      \rowcolor[gray]{0.9} \textbf{w.} \ralign{}
      & 87.53$_{\uparrow2.07}$ 
      & 87.33$_{\uparrow2.30}$  
      & 85.86$_{\uparrow1.73}$ 
      \\

      \midrule

      Janus-Pro-7B
      & 88.41
      & 87.30 
      & 85.70 
      \\

      \rowcolor[gray]{0.9} \textbf{w.} \ralign{}
      & 89.73$_{\uparrow1.32}$ 
      & 88.37$_{\uparrow1.07}$
      & 86.27$_{\uparrow0.57}$
      \\

      \midrule

      \makecell[l]{Qwen2.5-VL- \\ \quad \quad 3B-Instruct}
      & 88.32	
      & 87.60	
      & 86.63
      \\

      \rowcolor[gray]{0.9} \textbf{w.} \ralign{}
      & 89.69$_{\uparrow1.37}$	
      & 88.33$_{\uparrow0.73}$	
      & 87.16$_{\uparrow0.53}$
      \\

      \midrule

      \makecell[l]{Qwen2.5-VL- \\ \quad \quad 7B-Instruct}
      & 88.73	
      & 87.90	
      & 86.87
      \\

      \rowcolor[gray]{0.9} \textbf{w.} \ralign{}
      & 89.27$_{\uparrow0.54}$	
      & 88.10$_{\uparrow0.20}$	
      & 87.10$_{\uparrow0.23}$
      \\

      \midrule

      LLaVA-v1.5-7B
      & 88.14 
      & 87.23
      & 85.10
      \\

      \textbf{w.} LLaVA-RLHF
      & 84.77$_{\downarrow3.37}$
      & 84.60$_{\downarrow2.63}$
      & 83.40$_{\downarrow0.50}$
      \\

      \textbf{w.} POVID
      & 88.21$_{\uparrow0.07}$
      & 87.16$_{\downarrow0.07}$
      & 85.06$_{\downarrow0.04}$
      \\

      \textbf{w.} CSR (3Iter)
      & 87.83$_{\downarrow0.31}$
      & 87.00$_{\downarrow0.23}$
      & 85.00$_{\downarrow0.10}$
      \\

      \textbf{w.} SIMA
      & 88.10$_{\downarrow0.04}$
      & 87.10$_{\downarrow0.13}$
      & 85.03$_{\downarrow0.07}$
      \\

      \textbf{w.} mDPO
      & 88.17$_{\uparrow0.03}$
      & 87.13$_{\downarrow0.10}$
      & 85.03$_{\downarrow0.07}$
      \\

      \rowcolor[gray]{0.9} \textbf{w.} \ralign{}
      & 88.65$_{\uparrow0.51}$
      & 87.43$_{\uparrow0.20}$
      & 85.16$_{\uparrow0.06}$
      \\

      \midrule

      LLaVA-v1.5-13B
      & 88.07
      & 87.53
      & 85.60
      \\

      \textbf{w.} CSR (3Iter)
      & 88.38$_{\uparrow0.31}$
      & 87.90$_{\uparrow0.37}$
      & 85.46$_{\downarrow0.14}$
      \\

      \textbf{w.} SIMA
      & 88.04$_{\downarrow0.03}$
      & 87.40$_{\downarrow0.13}$
      & 85.40$_{\downarrow0.20}$
      \\

      \textbf{w.} HSA-DPO
      & 85.01$_{\downarrow3.06}$
      & 85.00$_{\downarrow2.53}$
      & 83.86$_{\downarrow1.74}$
      \\

      \rowcolor[gray]{0.9} \textbf{w.} \ralign{}
      & 90.03$_{\uparrow1.96}$
      & 89.20$_{\uparrow1.30}$
      & 86.20$_{\uparrow0.74}$
      \\

      \midrule

      \makecell[l]{LLaVA-v1.6-\\ \quad \quad Vicuna-7B}
      & 88.52
      & 87.63
      & 86.36
      \\

      \rowcolor[gray]{0.9} \textbf{w.} \ralign{}
      & 88.94$_{\uparrow0.42}$ 
      & 88.03$_{\uparrow0.40}$ 
      & 86.63$_{\uparrow0.27}$ 
      \\

      \midrule

      \makecell[l]{LLaVA-v1.6-\\ \quad \quad Vicuna-13B}
      & 88.24
      & 87.70
      & 86.43
      \\

      \rowcolor[gray]{0.9} \textbf{w.} \ralign{}
      & 88.79$_{\uparrow0.55}$ 
      & 88.10$_{\uparrow0.40}$ 
      & 86.60$_{\uparrow0.17}$ 
      \\
      
      \bottomrule
    \end{tabular}
  \end{center}
  \caption{Impact of \ralign{} across various model scales on POPE.}
  \label{tab:scale-pope}
\end{table}
\paragraph{Results}
Table \ref{tab:scale-pope} presents the performance of \ralign{} using both standard image-to-text and unified VLM backbones across model sizes from 1B to 13B on the POPE benchmark~\citep{pope}. In experiments with the LLaVA-v1.5 series~\citep{liu2024llava}, none of the baseline approaches consistently improve performance for either the 7B or the 13B models, highlighting the limited scalability of these methods. In contrast, \ralign{} achieved substantial performance gains, outperforming both the baseline models and the vanilla version—most notably on the LLaVA-v1.5-13B variant. Similarly, experiments with the LLaVA-v1.6-Vicuna series~\citep{llavanext} and Qwen2.5-VL series~\citep{bai2025qwen2.5vl} revealed the same trend, further underscoring \ralign{}'s superior scalability.
For unified vision-language models, especially Janus-Pro, integrating \ralign{} yields a significant performance boost. Notably, Janus-Pro-1B experiences the greatest improvement, underscoring \ralign{}’s robustness across different model architectures. However, Janus-Pro-1B, being the smallest among the evaluated VLMs, also exhibits the poorest overall performance on POPE, suggesting a correlation between model size and the propensity for hallucinations.

\subsection{Ablation Study}
In this section, we conduct a comprehensive ablation study to explore how the data curation framework and design of the objective function affect the \ralign{}' performance. The experimental setup follows the same setting as VLMs alignment experiments, with LLaVA-1.5-7B as the backbone. 

\paragraph{Dataset} Due to budget constraints and the need for reproducibility, we have excluded benchmarks that require evaluation by GPT-4~\citep{gpt4}. Instead, we focus on the following tasks: ScienceQA~\citep{lu2022scienqa}, TextVQA~\citep{singh2019towards}, and POPE~\citep{pope}.


\begin{table}[h]
  \footnotesize
  \setlength{\tabcolsep}{5pt}
  \begin{center}
    \begin{tabular}{lcccccccccccccccccccr}
      \toprule

      $\tau$
      & SQA
      & TextVQA
      & POPE$^r$
      & POPE$^p$
      & POPE$^a$
      \\
      \midrule

      \textit{\textbf{0.85}}
      & 67.04
      & 57.31
      & 88.96
      & 87.83
      & 85.06
      
      \\

      \textbf{\textit{0.90}}
      & 67.75
      & 57.68
      & 88.83
      & 87.66
      & 84.93
      
      \\

      \rowcolor[gray]{0.9}  \textbf{\textit{0.95}}
      & 68.10
      & 58.55
      & 88.65
      & 87.43
      & 85.16
      \\

      \bottomrule
    \end{tabular}
  \end{center}
  \caption{Impact of similarity threshold $\tau$ for generating the rejected responses in \ralign{} across general and hallucination benchmarks for VLMs.}
  \label{tab:tau-effect}
\end{table}

\paragraph{Similarity Threshold $\tau$}
In \ralign{}, we set the similarity threshold $\tau$ to 0.95, which acts as an upper bound on the cosine similarity between the chosen response and the generated rejected response. As illustrated in Table~\ref{tab:tau-effect}, decreasing the threshold $\tau$ results in a stronger preference signal, leading to improved performance in mitigating hallucinations. However, this comes at the cost of reduced performance in general VQA. Among the evaluated configurations, setting $\tau = 0.95$ offers the best trade-off, effectively reducing hallucinations while maintaining strong performance across VQA benchmarks.

\paragraph{Masking Strategy} In data curation, we generate preference data by inducing hallucinations at the segment level. To further investigate the impact of finer-grained perturbations, we conduct experiments using sentence-level masking. As shown in Table \ref{tab:mask-effect}, using a sentence-level masking strategy, \ralign{} still demonstrates significant improvement in reducing hallucination in VLMs. However, this approach leads to a slight drop in performance on general VQA tasks. More discussions on the masking strategy can be found in Appendix \ref{app:add-dis}.

\begin{table}[htbp]
  \footnotesize
  \setlength{\tabcolsep}{3pt}
  \begin{center}
    \begin{tabular}{lcccccccccccccccccccr}
      \toprule

      \makecell{Masking\\Strategy}
      & SQA
      & TextVQA
      & POPE$^r$
      & POPE$^p$
      & POPE$^a$
      \\
      \midrule

      \textbf{\textit{sentence-level}}
      & 67.58
      & 57.77
      & 88.56
      & 87.60
      & 84.90
      \\

      \rowcolor[gray]{0.9} \textit{\textbf{segment-level}}
      & 68.10
      & 58.55
      & 88.65
      & 87.43
      & 85.16
      \\

      \bottomrule
    \end{tabular}
  \end{center}
  \caption{Impact of masking strategy across general and hallucination benchmarks for VLMs.}
  \label{tab:mask-effect}
\end{table}


\paragraph{Design of Loss Function} In \ralign{}, we assign equal weights to the DPO and vDPO objectives in the combined loss function, i.e., $\mathcal{L}_{\text{rDPO}} = \mathcal{L}_{\text{DPO}} + \mathcal{L}_{\text{vDPO}}$. To better understand the impact of this design of loss function, we generalize the loss function to $\mathcal{L}_{\text{DPO}} + w_v \mathcal{L}_{\text{vDPO}}$, where $w_v$ controls the contribution of the visual component, and conduct experiments with different values of $w_v$ to analyze the trade-offs and identify the optimal balance between textual and visual preference signals. As shown in Table~\ref{tab:rdpo-effect}, incorporating the $\mathcal{L}_{\text{vDPO}}$ objective significantly enhances VLM performance on hallucination benchmarks. In general, when combined with the standard $\mathcal{L}_{\text{DPO}}$ objective, increasing the weight of $\mathcal{L}_{\text{vDPO}}$ tends to yield better overall performance. Notably, the equally-combined objective $\mathcal{L}_{\text{rDPO}}$ achieves the best balance between reducing hallucinations and maintaining strong performance on general VQA benchmarks, highlighting its effectiveness as a robust training strategy.

\begin{table}[htbp]
  \footnotesize
  \setlength{\tabcolsep}{3pt}
  \begin{center}
    \begin{tabular}{lcccccccccccccccccccr}
      \toprule

      $w_v$
      & SQA
      & TextVQA
      & POPE$^r$
      & POPE$^p$
      & POPE$^a$
      \\
      \midrule


      \textbf{\textit{0.0}} (DPO)
      & 66.26
      & 58.24
      & 88.18
      & 87.30
      & 85.23
      \\

      

      \textbf{\textit{0.25}}
      & 67.15
      & 57.47
      & 88.72
      & 87.60	
      & 85.03

      \\

      \textbf{\textit{0.50}}
      & 67.01
      & 57.41
      & 88.76
      & 87.53
      & 85.06

      \\

      \textbf{\textit{0.75}}
      & 67.53
      & 57.69
      & 88.90
      & 87.70
      & 84.83

      \\

      \rowcolor[gray]{0.9} \textbf{\textit{1.0}} (rDPO)
      & 68.10
      & 58.55
      & 88.65
      & 87.43
      & 85.16
      \\

      \bottomrule
    \end{tabular}
  \end{center}
  \caption{Impact of rDPO objective across general and hallucination benchmarks for VLMs, and comparisons with baselines.}
  \label{tab:rdpo-effect}
\end{table}

\paragraph{Training Epochs}

For a fair comparison with prior baselines, we primarily report results of \ralign{} under a one-epoch fine-tuning setup, which already demonstrates the effectiveness of our proposed method. To further explore the impact of training duration, we conduct additional experiments with extended fine-tuning of up to three epochs.

\begin{table}[htbp]
  \footnotesize
  \setlength{\tabcolsep}{3pt}
  \begin{center}
    \begin{tabular}{lcccccccccccccccccccr}
      \toprule

      \makecell{Num\\Epoch}
      & SQA
      & TextVQA
      & POPE$^r$
      & POPE$^p$
      & POPE$^a$
      \\
      \midrule

      \textbf{1}
      & 68.10	
      & 58.55	
      & 88.65	
      & 87.43	
      & 85.16
      \\

      \textbf{2}
      & 68.27	
      & 58.47	
      & 88.91	
      & 87.52	
      & 85.16
      \\

      \textbf{3}
      & 68.17	
      & 58.60	
      & 88.57	
      & 87.60	
      & 85.43
      \\

      \bottomrule
    \end{tabular}
  \end{center}
  \caption{Impact of the number of training epochs across general and hallucination benchmarks for VLMs.}
  \label{tab:epoch-effect}
\end{table}

As shown in Table~\ref{tab:epoch-effect}, \ralign{} exhibits stable performance across longer training schedules, with results consistently maintained and in some cases slightly improved on both general VQA benchmarks (SQA, TextVQA) and hallucination benchmarks (POPE). This indicates that our method is robust to extended training and not prone to overfitting, while continuing to deliver reliable gains.

\section{Discussions} 

\paragraph{Role of Image $v_l$}

$v_l$ is one of the top-10 retrieved images corresponding to the original image $v$, and qualitatively, the images $v$ and $v_l$ 
are semantically similar in terms of scenes, objects, and composition. This retrieval strategy is intended to ensure that 
$v_l$ shares sufficient visual context with $v$, making it a plausible alternative grounding for the instruction $x$. Furthermore, we compute the cosine similarity between the CLIP embeddings of the caption of $v$ (by prompting "Describe this image in detail.") and three types of images: the original image  $v$, a retrieved image $v_l$, and a randomly selected image $v_r$. The average cosine similarities are $0.2780, 0.2382, 0.0688$, respectively, which indicates that $v_l$ retains significant semantic similarity with $v$ and is far more aligned than an unrelated image $v_r$. Based on this, we interpret $v_l$ as a ``\textit{rejected input image}'' to the original instruction $x$: it provides a visually plausible but suboptimal context, under which the response $y_w$ should be less preferred compared to when conditioned on $v$.

\paragraph{Discussion with mDPO} In this section, we detail the differences between our proposed rDPO and mDPO~\citep{wang2024mdpo}. In mDPO, a conditional preference optimization  objective is introduced to force the model to determine the preference label based on visual information:
\begin{align*}
    &\mathcal{L}_{\text{CoDPO}} = - \mathbb{E}_{(x,v,y_w,y_l) \sim \mathcal{D}}\\
    &\bigg[\log \sigma \bigg(\beta \log \frac{\pi_\theta (y_w|x,v)}{\pi_0 (y_w|x,v)}
    - \beta \log \frac{\pi_\theta (y_w|x,v_c)}{\pi_0 (y_w|x,v_c)} \bigg) \bigg],
\end{align*}
where $v_c$ denotes a randomly cropped image of the original input image $v$. Specifically, visual preference signals are generated by randomly masking $20\%$ of the input visual tokens to encourage the model to capture preferences based on visual cues.

In contrast, \ralign{} extends and enhances this approach by incorporating a more semantically meaningful visual preference pair. Instead of relying solely on random crops, \ralign{} retrieves a relevant image from the same dataset that corresponds to the original input. This retrieval-based augmentation provides a stronger contrastive signal, improving the model’s ability to discern fine-grained visual details and reducing spurious correlations. Moreover, beyond mitigating hallucinations in VLMs, \ralign{} has been demonstrated that it also significantly enhance performance on general VQA tasks.









\paragraph{Performance Variations on General VQA tasks}

While \ralign{} consistently delivers the best performance on hallucination benchmarks across all backbone models, it may not achieve the top result for every general VQA benchmark. The variations in performance on general VQA tasks are primarily due to the alignment tax, a well-known phenomenon in RLHF, where alignment can sometimes lead to a decline in the model’s ability to retain pretraining knowledge. Notably, this trade-off is not unique to \ralign{}; as shown in Table \ref{tab:general-task}, several baselines even underperform compared to the vanilla VLMs on general VQA tasks.

\label{app:add-dis}
\paragraph{Segment-level Preference} 
\begin{figure}[htbp]
    \centering
    \includegraphics[width=.7\linewidth]{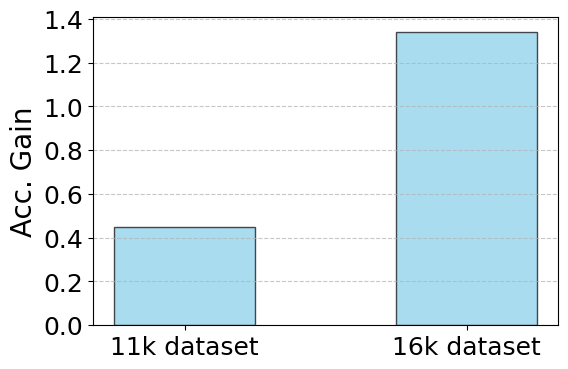}
    \caption{Performance gains of \ralign{} with LLaVA-v1.6-Mistral-7B  as the backbone on ScienceQA with respect to the size of preference data.}
    \label{fig:sqa-size}   
\end{figure}
Building on the findings of \cite{yu2024rlhf}, we generate preference data by inducing hallucinations at the segment level rather than at the sentence level (as seen in approaches such as POVID~\citep{zhou2024povid}, STIC~\citep{deng2024stic}, and CSR~\citep{zhou2024calibrated}), to provide robust supervision signals during the alignment process. This finer-grained preference modeling yields clearer and more precise learning signals, enabling the model to better distinguish between subtle hallucinations and ground truth responses. To further investigate these segment-level preference signals, we expanded the fine-tuning dataset from $11k$ to $16k$ image samples. As illustrated in Figure \ref{fig:sqa-size}, when using LLaVA-v1.6-Mistral-7B as the backbone with ScienceQA as the case study, \ralign{} achieved a significant performance improvement—from $0.45$ to $1.34$—demonstrating the effectiveness of our approach.

\paragraph{Computational Complexity}

The proposed \ralign{} pipeline can be modularized into offline preprocessing and online training integration (detailed computational cost can be found in the Appendix):

\begin{itemize}[leftmargin=*]
    \item \textbf{Preprocessing:} Image retrieval, strategic masking, and preference pair generation can be entirely performed offline as a one-time data preprocessing step. This includes CLIP-based similarity search, mask generation, and SentenceTransformer-based similarity computation. Once completed, these preprocessed preference pairs can be reused across multiple training runs without additional overhead.

    \item \textbf{Training Overhead:} The actual training process introduces minimal additional computational overhead (~5-10\% increased training time) compared to standard DPO, with virtually identical memory requirements. The additional cost stems only from: 
    \begin{itemize}[leftmargin=1em]
        \item Forward passes through the visual encoder for retrieved images;
        \item Generation passes through the LLM backbone for computing the vDPO loss component.
    \end{itemize}
\end{itemize}

\section{Related Work}

\paragraph{Reinforcement Learning from Human Feedback} 
Reinforcement Learning from Human Feedback (RLHF) has emerged as a crucial technique for incorporating human preference signals into machine learning methods and models~\citep{dong2024rlhf,yin2022offline}. RLHF frameworks can be broadly categorized into deep RL-based approaches and direct preference learning approaches. In deep RL-based methods, a reward model is first constructed, after which Proximal Policy Optimization (PPO)~\citep{schulman2017proximal, christiano2017deep, ziegler2019fine} is employed to optimize the reward signals with KL regularization~\citep{ouyang2022training, touvron2023llama2}. While the direct preference learning approaches optimize a designed loss target on the offline preference dataset directly, eliminating the need for a separate reward model~\citep{rafailov2024direct,ipo,gpo,ethayarajh2024kto}.

\paragraph{Vision Language Models} 
Large Vision Language Models (VLMs)~\citep{li2022blip, li2023blip2, liu2024llava,llavanext,llama3.2, Qwen-VL, Qwen2VL, lu2024deepseek, wu2024deepseek,bai2025qwen2.5vl,fan2025vlm3rvisionlanguagemodelsaugmented,abouelenin2025phi4-mini} extended the understanding and reasoning capabilities of Large Language Models (LLMs)~\citep{devlin2018bert, radford2019gpt2,brown2020gpt3,team2023gemini,roziere2023codellama,touvron2023llama,touvron2023llama2, raffel2020t5,qwen2,qwen2.5,pan2024plum,yang2025qwen3} into the visual domain. By integrating vision encoders, such as CLIP~\citep{radford2021clip}, image patches are first converted into embeddings and then projected to align with text embedding space, unlocking unprecedented cross-modal applications in the real world, such as biomedical imaging~\citep{moor2023med,li2024llava-med,zuo20254kagent}, autonomous systems~\citep{shao2024lmdrive,tian2024drivevlm,sima2023drivelm,openemma, ma2025position,wang2025generative,li2025mmhu,gao2025langcoop}, and robotics~\citep{rana2023sayplan,kim2024openvla, xing2025can}.

\paragraph{Alignment of Vision Language Models}
Current VLMs often suffer from hallucinations, producing inaccurate or misleading information that fails to accurately represent the content of the provided image~\citep{zhu2024unraveling,bai2024hallucination,qian2025decalign,xing2025demystifying}. Such misalignments can have catastrophic consequences when these models are deployed in real-world scenarios~\citep{autotrust}.
To address cross-modality hallucinations, recent research has primarily focused on applying direct preference optimization~\citep{deng2024stic,zhou2024povid,fang2024vila,zhou2024calibrated,guo2024direct,chen2024dress,wang2024enhancing,yu2024rlhf,li2023silkie,wang2024mdpo} or contrastive learning~\citep{sarkar2024mitigating} on the curated datasets with preference signals, and utilizing model editing techniques~\citep{liu2024paying,yu2024attention}.

\section{Conclusion}
In this paper, a novel framework, \ralign{}, for aligning VLMs to mitigate hallucinations is proposed. Our approach leverages image retrieval to deliberately induce segment-level hallucinations, thereby generating plausible and natural preference signals. By integrating the retrieved images, a dual-preference dataset that encompasses both textual and visual cues is curated. Furthermore, we propose the rDPO objective, an extension of DPO that includes an additional visual preference optimization objective, to enhance the alignment process with valuable visual preference signals. Comprehensive empirical results from a range of general VQA and hallucination benchmarks demonstrate that \ralign{} effectively reduces hallucinations in VLMs while enhancing their overall performance. Moreover, it demonstrates superior scalability across various model architectures and sizes.

\section*{Limitations} 

Although \ralign{} has demonstrated superior performance on both hallucination and general VQA benchmarks, it does not always achieve state-of-the-art results on general tasks; in some cases, its performance is even worse than that of vanilla VLMs. Future research could explore strategies to eliminate this alignment tax or identify an optimal balance for this trade-off. 

The potential risks of this work align with the general challenges of RLHF alignment. As more powerful alignment techniques are developed, they may inadvertently empower adversarial approaches that exploit these models, potentially leading to unfair or discriminatory outputs. Meanwhile, these adversarial strategies can be used to generate negative samples, which can ultimately contribute to the development of more robust and reliable VLMs.

\clearpage
\bibliography{main}

\appendix
\clearpage

\begin{table*}[htbp]
  \footnotesize
  \setlength{\tabcolsep}{3.5pt}
  \begin{center}
  \renewcommand{\arraystretch}{1.5}
    \begin{tabular}{lccccc}
      \toprule
      \textbf{Methods} & \textbf{Source} & \textbf{Size} & \textbf{Preference Signal} & \textbf{Curation Strategy} & \textbf{Visual Modification} \\
      \midrule
      LLaVA-RLHF & LLaVA-Instruct & 10k & Textual only & Human annotation & None \\
      \rowcolor[gray]{0.9}POVID      & LLaVA-Instruct & 17k & Textual only & Image noising + prompting & Gaussian noise \\
      CSR        & LLaVA-Instruct & 13k & Textual only & Self-rewarding & None \\
      \rowcolor[gray]{0.9}SIMA       & COCO           & 5k  & Textual only & Self-rewarding & None \\
      STIC       & COCO           & 6k  & Textual only & Cropping Image + prompting & \makecell[c]{Color jitter +\\ lower resolution} \\
      \rowcolor[gray]{0.9}\textbf{Re-Align} & LLaVA-Instruct & 11k & Textual \& Visual & Image retrieval + strategic masking & \makecell[c]{Semantically-guided \\natural images} \\
      \bottomrule
    \end{tabular}
  \end{center}
  \caption{Summary of preference datasets used in \ralign{} and baseline methods. Dataset sizes reflect only preference pairs used for alignment training, not the total datasets involved in each method. Several baselines additionally rely on larger supervised fine-tuning datasets.}
  \label{tab:pref-datasets}
\end{table*}

\section{Overview of \ralign{}}\label{app:alg-ralign}

\begin{algorithm}[]
\caption{Overview of \ralign{} }
\label{alg:harp}
\textbf{Required:}
\\ (1) Unlabeled images $\{v_i\}$ with instructions $\{x_i\}$;
\\ (2) an advanced VLM model $\mathcal{V}$;
\\ (3) caption masking prompt $P_m$;
\\ (4) masked caption completion prompt $P_c$;
\\ (5) a text encoder $\mathcal{T}$.

\textbf{Input:} A reference model $\pi_0$ with vision encoder $f_v(\cdot)$, VLM $\pi_\theta$, hyper-parameter $k, \tau$.
\begin{algorithmic}[1]
    \State $\mathcal{D} \gets \emptyset$  \textcolor{cvprblue}{// Init preference dataset}
    \State $N \gets |\{v_i\}|$
    \For{$i= 1, \cdots, N$}
    \State $y_w \gets \mathcal{V}(x_i, v_i)$ \textcolor{cvprblue}{// Get preferred response}
    \State $y_m \gets \mathcal{V}(P_m, x_i, v_i)$ \textcolor{cvprblue}{// Strategic masking}
    \State $s^j_i = \text{sim} (f_v(v_i), f_v(v_j)), \forall i\neq j$
    \State \textcolor{cvprblue}{// Retrieve top-$k$ similar images}
    \State $s^{j_1}_i, \cdots, s^{j_k}_i \gets \text{Top}_k(s_i^j)$
    \State $y_l \gets \text{None}, v_l \gets \text{None}$
    \For{$t= 1, \cdots, k$} 
    \State \textcolor{cvprblue}{// Generate candidate hallucinations}
    \State $y_c \gets \mathcal{V}(P_c, y_m, v_{j_t})$
    \If{sim$(\mathcal{T}(y_w), \mathcal{T}(y_c)) \geq \tau$}
    \State \textcolor{cvprblue}{// Assign rejected response}
    \State $y_l \gets y_c, v_l \gets v_{j_k}$
    \EndIf
    \EndFor
    \If{$y_l$ is None}
    \State \textbf{continue}
    \EndIf
    \State $\mathcal{D} \gets \mathcal{D} \cup \{x_i, v_i, v_l, y_w, y_l\}$
    \EndFor
    \State Update $\pi_\theta$ through $\mathcal{L}_{\text{rDPO}}$ (\cref{eq:rdpo})
    \State \Return $\pi_\theta$
\end{algorithmic}
\end{algorithm}

\section{Details of the Evaluated Baselines} We compare our proposed method with the following alignment frameworks for VLMs:

\begin{itemize}[leftmargin=*]
    \item \textbf{LLaVA-RLHF}~\citep{sun2023aligning}: conducts SFT on for updating the projector only and then PPO on the preference data collected from human annotators.
    
    \item \textbf{POVID}~\cite{zhou2024povid}: constructing preference data by prompting GPT-4V~\citep{gpt4v} to generate hallucinations while intentionally injecting noise into image inputs, followed by fine-tuning VLMs using DPO.

    \item \textbf{CSR}~\citep{zhou2024calibrated}: iteratively generates candidate responses and curates preference data using a self-rewarding mechanism, followed by fine-tuning VLMs via DPO. 

    \item \textbf{SIMA}~\citep{wang2024enhancing}: self-generates responses and employs an in-context self-critic mechanism to select response pairs for preference data construction, followed by fine-tuning with DPO.

    \item \textbf{STIC}~\citep{deng2024stic}: self-generates chosen responses and constructs preference data by introducing corrupted images or misleading prompts, followed by fine-tuning with regularized DPO. 

    \item \textbf{mDPO}~\citep{wang2024mdpo}: finetunes the model with conditional preference optimization, which incorporates an additional objective to account for image-level preferences and a reward anchor that forces the reward to be positive for chosen responses.
\end{itemize}

\section{Prompts used for Preference Data Construction}

During the construction of the preference dataset for \ralign{}, we employed GPT-4o mini~\citep{gpt4omini} to mask the chosen response using the following prompt.

\begin{tcolorbox}[colback=gray!5!white, colframe=gray!75!black, 
title=Strategic Masking]
        Please mask any words of the segments related to the objects, attributes, and logical relationships of the input image in the following description by replacing them with [MASK].\\
\end{tcolorbox}

Then, we instruct the VLMs to produce a candidate completion for the masked response to generate the final rejected response using the following prompt.

\begin{tcolorbox}[colback=gray!5!white, colframe=gray!75!black, 
title=Masking Completion]
        Please complete the following sentence based on the input image by filling in the masked segments.\\
\end{tcolorbox}

\section{Examples of Preference Pair}

\begin{figure}[h]
    \centering
    \includegraphics[width=1.\linewidth]{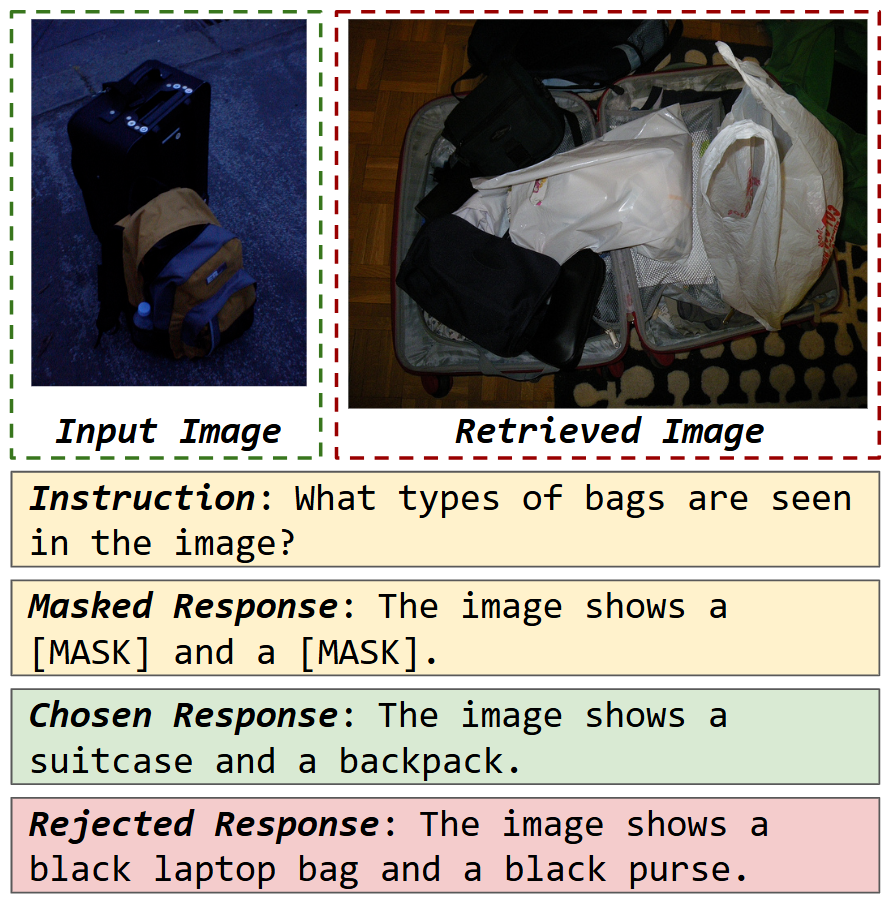}
    \caption{Example preference pair for VQA generated using \ralign{}.}
    \label{fig:vqa-case} 
\end{figure}

\begin{figure*}[htbp]
    \centering
    \includegraphics[width=0.9\linewidth]{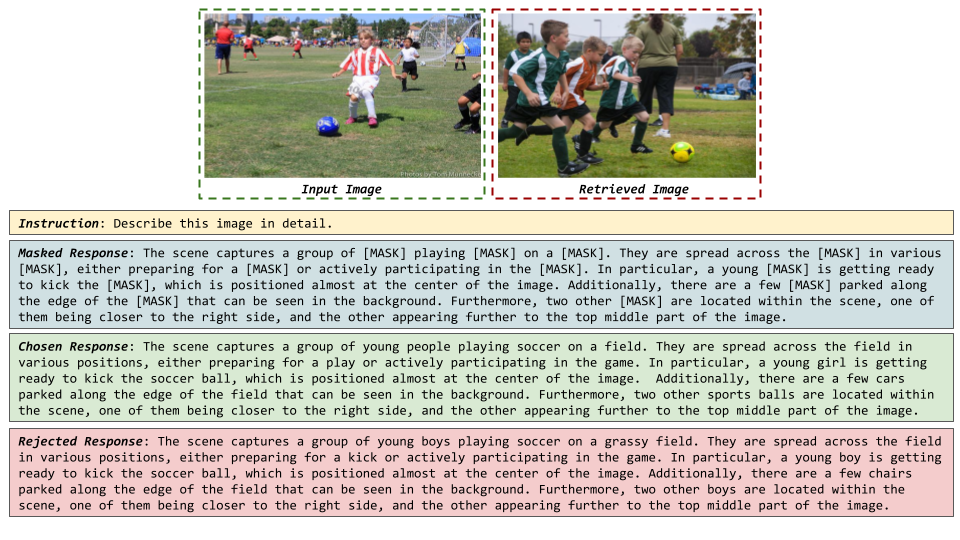}
    \caption{Example preference pair for image captioning generated using \ralign{}.}
    \label{fig:iamge-cap-case} 
\end{figure*}

Table \ref{fig:vqa-case} and \ref{fig:iamge-cap-case} provide examples of the constructed preference data for the VQA and image captioning, and each data sample contains textual instruction, input image, retrieved image, chosen response, and rejected response.

\begin{figure*}[h]
    \centering
    \includegraphics[width=0.9\linewidth]{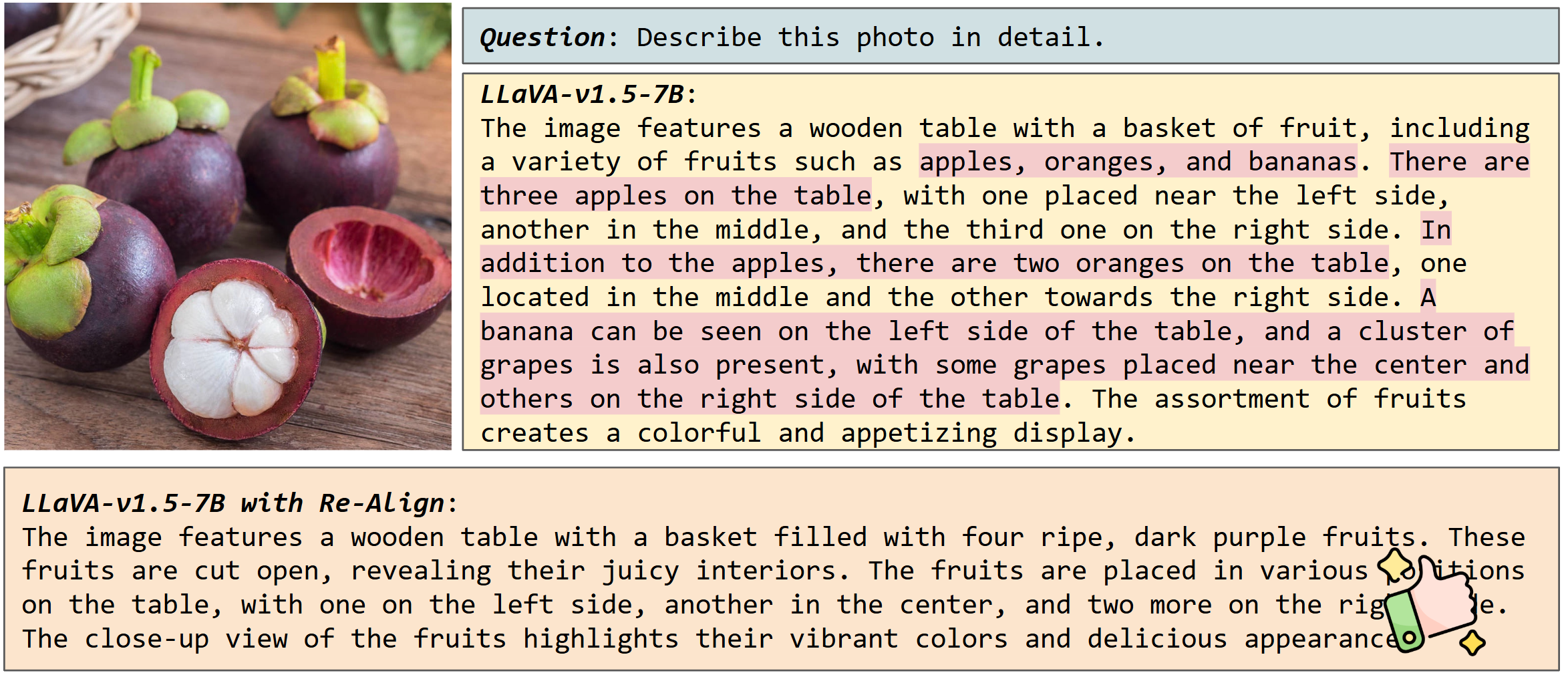}
    \caption{Example responses generated by LLaVA-v1.5-7B and \ralign{}.}
    \label{fig:response-case} 
\end{figure*}

\section{Response Examples} 
Figure \ref{fig:response-case} presents example responses from both the original LLaVA-v1.5-7B model and \ralign{} as evaluated on LLaVABench. Notably, the original model's response exhibits server object hallucinations, while \ralign{} delivers a clearer and more accurate description of the image.

\section{Data Curation}
\label{appendix:pref-data}

Table~\ref{tab:pref-datasets} summarizes the key characteristics of the preference datasets employed by \ralign{} and several baseline alignment methods. Importantly, the reported dataset sizes correspond only to the preference pairs used directly for alignment training, and not to the total datasets leveraged in each pipeline. Several baseline methods, such as LLaVA-RLHF and POVID, additionally rely on larger supervised fine-tuning stages with external datasets, whereas \ralign{} operates solely on curated preference data.  

Unlike baselines that depend on synthetic perturbations or expensive human annotations, \ralign{} introduces a semantically-guided image retrieval and masking procedure to construct preference datasets. This strategy offers several critical advantages:  

\begin{itemize}[leftmargin=*]
  \item \textbf{Semantic Coherence.} Retrieved natural images preserve contextual integrity and semantic relationships, which are often degraded by cropped or artificially edited images. 
  \item \textbf{Natural Preference Signals.} The curated pairs reflect genuine visual understanding rather than superficial low-level perturbations (e.g., Gaussian noise, color jitter, or downsampling artifacts).    
\end{itemize}  

The construction of preference data is a key determinant of downstream alignment performance. By leveraging semantically-guided retrieval, \ralign{} produces preference pairs that are both semantically rich and visually natural, contributing to its robustness across both general VQA and hallucination benchmarks.

\section{Licenses}
\label{sec:licenses}

The LLaVA-Instruct-150K dataset~\citep{liu2024llava} which is used to construct preference data is released under CC BY 4.0 license and it should abide by the policy of OpenAI\footnote{https://openai.com/policies/terms-of-use}.

For the hallucination benchmarks, POPE~\citep{pope} and HallusionBench~\citep{guan2023hallusionbench} are released under MIT and BSD-3-Clause licenses. 

For the general VQA benchmarks, ScienceQA~\citep{lu2022scienqa}, TextVQA~\citep{singh2019towards}, MM-Vet~\citep{yu2023mm}, VisWiz~\citep{gurari2018vizwiz}, LLaVABench~\citep{llavabench}, and MMBench~\citep{liu2024mmbench} are released under MIT, CC BY 4.0, Apache-2.0, CC BY 4.0, Apache-2.0, and Apache-2.0 licenses respectively. While MME~\citep{Fu2023mme} was released without an accompanying license.

\section{Experimental Cost}

The cost for curating the preference dataset by using GPT-4o mini~\citep{gpt4omini} cost approximately \$90 in total.The evaluation of HallusionBench and LLaVABench using GPT-4~\citep{gpt4} incurred an approximate total cost of \$30.

\section{Computational Cost} 
All fine-tuning and evaluation experiments were executed on four NVIDIA A6000ada GPUs. Table \ref{tab:time} details the time required for \ralign{} to fine-tune each model.

\begin{table}[htbp]
  \footnotesize
  \begin{center}
    \begin{tabular}{lc}
      \toprule

      Models
      & Required Time
      \\
      \midrule

      Janus-Pro-1B
      & 50 min
      
      \\

      Janus-Pro-7B
      & 93 min
      
      \\

      LLaVA-v1.5-7B
      & 35 min
     
      \\

      LLaVA-v1.5-13B
      & 45 min
      
      \\

      LLaVA-v1.6-Mistral-7B
      & 30 min
      \\
      
      \makecell[l]{LLaVA-v1.6-Vicuna-7B}
      & 46 min
      
      \\

      \makecell[l]{LLaVA-v1.6- Vicuna-13B}
      & 72 min
      \\
      
      \bottomrule
    \end{tabular}
  \end{center}
  \caption{Time required for fine-tuning VLMs with \ralign{}.}
  \label{tab:time}
\end{table}

\section{Hyperparameter Setting} 

For all the experiments, we fine-tuning VLMs with \ralign{} for 1 epoch. We deploy LoRA fine-tuning with \texttt{lora\_r}=128, \texttt{lora\_alpha}=256, \texttt{target\_module}=all, and hyperparameters as presented in Table \ref{tab:hypeterparameter}. 

\begin{table}[htbp]
  \footnotesize
  \begin{center}
    \begin{tabular}{ll}
      \toprule

      Hyperparameter
      & Setting 
      \\
      \midrule

      $\beta$
      & 0.1
      \\

      Learning rate
      & 1e-5
      \\

      \texttt{weight\_decay}
      & 0.0
      \\

      \texttt{warmup\_ratio}
      & 0.03
      \\

      \texttt{lr\_scheduler\_type}
      & \texttt{cosine}
      \\

      \texttt{mm\_projector\_lr}
      & 2e-5
      \\

      \texttt{mm\_projector\_type}
      & mlp2x\_gelu
      \\

      \texttt{gradient\_accumulation\_steps}
      & 8 
      \\

      \texttt{per\_device\_train\_batch\_size}
      & 1
      \\

      \texttt{bf16}
      & True
      \\

      Optimizer
      & AdamW
      
      \\
      
      \bottomrule
    \end{tabular}
  \end{center}
  \caption{Hypeterparameter setting for fine-tuning.}
  \label{tab:hypeterparameter} 
\end{table}

\section{ Social Impacts}
Our proposed novel alignment framework for VLMs, \ralign{}, not only significantly mitigates the hallucinations of VLMs but also elevates their generalization capabilities across diverse multimodal tasks. These advancements hold far-reaching societal implications, particularly in advancing the development of trustworthy, ethically aligned AI systems capable of reliable real-world deployment. To elucidate these implications, we provide a comprehensive overview of potential transformative outcomes:
\begin{itemize}[leftmargin=*]
    \item \textbf{Enhancing trustworthiness:} \ralign{} significantly enhances the reliability of AI-generated content by reducing hallucinated outputs and improving factual grounding. This ensures that users and regulatory bodies can place increased confidence in AI-driven decisions and recommendations.

    \item \textbf{Safety-critical applications:} By reducing erratic outputs and improving contextual awareness, \ralign{} enables safer deployment of VLMs in high-stakes domains such as healthcare diagnostics, autonomous vehicles, and disaster response systems, where error margins are near-zero and algorithmic trust is paramount.

    \item \textbf{Democratizing access to robust AI:} Our method can democratize access to advanced multimodal AI models under low-resource or data-scarce settings, which empowers researchers and practitioners with limited computational resources to participate in cutting-edge AI development, ultimately contributing to a more equitable and diverse AI ecosystem.
\end{itemize}

\section{Broader Impacts}

The research presented in this paper, particularly the development of the Re-Align framework, has significant broader impacts that extend beyond the immediate technical contributions. By improving the alignment of Vision Language Models (VLMs), our work contributes to the creation of more reliable, trustworthy, and capable AI systems, which have profound implications for various societal domains.

A primary impact of this research is the enhancement of safety and trustworthiness in AI systems deployed in critical applications. The reduction of hallucinations is paramount for autonomous systems where perception and decision-making must be grounded in reality. For instance, in autonomous driving, reliable visual understanding is non-negotiable. Our work aligns with efforts to build end-to-end autonomous driving models \citep{openemma,luo2025v2x}, improve motion prediction through equivariant geometry \citep{wang2023equivariant, wang2023eqdrive}, and multi-agent communication \citep{wang2025cmp, wang2025uniocc}. By ensuring that a VLM's outputs are faithful to its visual inputs, Re-Align contributes to the foundational safety required for deploying these technologies. The principles extend to other domains like robotics and collaborative agent systems, where trustworthy AI is essential for safe and effective operation \citep{li2025safeflow, gao2025airv2x, chen2024fair}.

Furthermore, our work contributes to the broader unification and advancement of generative and discriminative AI models. The alignment techniques we propose are part of a larger trend towards creating more cohesive and capable foundation models \citep{liu2024toward}. This advancement enables a wide range of new applications. For example, improved visual fidelity is crucial for tasks like novel view synthesis from single RGBD images \citep{hetang2023novel} and for understanding complex 3D environments from partial data \citep{zhang2021point}. As these models become more robust, they can be applied to creative industries, virtual reality, and scientific visualization with greater confidence.

Finally, the development of more effective and efficient alignment techniques has implications for the accessibility and democratization of AI. As methods like Direct Preference Optimization (DPO) become more refined, they can potentially lower the barrier to fine-tuning powerful models for specific, beneficial purposes. Techniques that improve the learning process, such as prompt learning using metaheuristics \citep{pan2024plum}, can make the customization of large models more efficient. However, it is crucial to acknowledge the dual-use nature of these powerful technologies. The same methods that align models to be helpful and harmless could potentially be used for malicious purposes. Therefore, ongoing research into robust safety protocols, ethical guidelines, and trustworthiness benchmarks \citep{autotrust} is essential to mitigate these risks and ensure that the societal benefits of advanced AI systems like those improved by Re-Align are realized responsibly.
\end{document}